\documentclass[letterpaper, 10 pt, journal, twoside]{ieeetran} 

\usepackage{color}
\usepackage{calc} 
\usepackage{hyperref}
\usepackage{aligned-overset}
\usepackage{cleveref}
\usepackage{graphicx}
\usepackage{floatflt} 
\usepackage{svg} 
\usepackage{comment}
\usepackage{siunitx}
\usepackage{tikz}
\usepackage{float}
\usepackage[shortlabels]{enumitem}
\usepackage[normalem]{ulem}
\usepackage{ wasysym }

\DeclareGraphicsExtensions{.pdf,.jpeg,.png,.pdf_tex,.eps}
\graphicspath{{./figure/}}

\IEEEoverridecommandlockouts    

\title{\LARGE \bf Environment-based Assistance Modulation for a Hip Exosuit via Computer Vision

\author{Enrica Tricomi$^{1}$, Mirko Mossini$^{2}$, Francesco Missiroli$^{1}$, Nicola Lotti$^{1}$, Michele Xiloyannis$^{3}$, \\ Loris Roveda$^{4}$, and Lorenzo Masia$^{1}$}

\thanks{$^{1}$ E. Tricomi, F. Missiroli, N. Lotti, and L. Masia are with the Institut f\"ur Technische Informatik (ZITI), Heidelberg University, 69120 Heidelberg, Deutschland.}

\thanks{$^{2}$ M. Mossini is with Department of Mechanical Engineering, Politecnico di Milano, via La Masa 1, 20156 Milan, Italy.}

\thanks{$^{3}$ M. Xiloyannis is with the Institute of Robotics and Intelligent Systems, ETH Z\"urich, Z\"urich, 8092, Switzerland. 
}%

\thanks{$^{4}$ L. Roveda is with Istituto Dalle Molle di studi sull'Intelligenza Artificiale (IDSIA), Scuola Universitaria Professionale della Svizzera Italiana (SUPSI), Università della Svizzera italiana (USI), via la Santa 1, 6962, Lugano, Switzerland}

\thanks{$*$ corresponding author: {\scriptsize\tt enrica.tricomi@ziti.uni-heidelberg.de}}

}

\begin{document}

\maketitle



\IEEEpeerreviewmaketitle

\begin{abstract}
Just like in humans vision plays a fundamental role in guiding adaptive locomotion, when designing the control strategy for a walking assistive technology, Computer Vision may bring substantial improvements when performing an environment-based assistance modulation.
In this work, we developed a hip exosuit controller able to distinguish among three different walking terrains through the use of an RGB camera and to adapt the assistance accordingly.
The system was tested with seven healthy participants walking throughout an overground path comprising of staircases and level ground.
Subjects performed the task with the exosuit disabled (\textit{Exo Off}), constant assistance profile (\textit{Vision Off}), and with assistance modulation (\textit{Vision On}).
Our results showed that the controller was able to promptly classify in real-time the path in front of the user with an overall accuracy per class above the $85\%$, and to perform assistance modulation accordingly.
Evaluation related to the effects on the user showed that \textit{Vision On} was able to outperform the other two conditions: we obtained significantly higher metabolic savings than \textit{Exo Off}, with a peak of $\approx$ $-20\%$ when climbing up the staircase and $\approx$ $-16\%$ in the overall path, and than \textit{Vision Off} when ascending or descending stairs.
Such advancements in the field may yield to a step forward for the exploitation of lightweight walking assistive technologies in real-life scenarios.
\end{abstract}


\section{Introduction}

Vision plays a fundamental role in bipedal walking, as it contributes in modulating navigation according to the surrounding space \cite{courtine2003human}.
In humans, environment scanning is a major actor in providing feedback loop to the central nervous system for cognitive-to-motor transformations, thereby endorsing adaptive locomotion \cite{koren2022vision}.

Within the context of human walking assistance through wearable devices, unlike humans, robot control strategies have been largely based on gait kinetics and kinematics, and investigations have often excluded the users' surrounding environment by studying locomotion in restricted cases (e.g., treadmill or straight overground walking) \cite{courtine2003human}. 
However, in the real world, it is very likely that users need to continuously and dynamically modify their natural gait pattern to adapt to stairs, ramps, or curved path \cite{kang2021real}.
To account for the change in kinematics when walking on dissimilar terrains, accurate modulation of the assistance delivered to the targeted joints, in terms of timing and magnitude, has to be provided to minimize walking energy expenditure and, therefore, to maximize users' benefits \cite{zhang2022enhancing, seo2016fully, quinlivan2017assistance, kim2022reducing}.

In the current state of the art, diverse methods addressed this issue implementing manual hand switching strategies or automatic techniques based on the user's biomechanics, yet identification and smooth transitions across different locomotion modalities remains one of the major challenges in assistive robotics \cite{qian2022predictive, tucker2015control, young2016state}.
Additional analyses about the user's extra-personal space should complement gait recognition strategies in order to have proper and tailored assistance modulation \cite{laschowski2021computer}.

As humans rely on vision and the underlying neural feedback to achieve this goal, a feasible approach to embed the aforementioned information in the control loop of robotic systems is to perform environment scanning and classification through computer vision.
The approach has been already extensively tested in the context of wheeled and legged robots \cite{fu2022coupling, kim2020vision}, where camera-based algorithms are developed in order to guide the system navigation on various terrains and path planning.
Nonetheless, a very small niche in literature is taken up by the inclusion of computer vision in the control of wearable assistive devices.
The idea is recently getting a foothold among researchers, but very few works exist so far \cite{laschowski2022environment, laschowski2021computer, quinlivan2017assistance, li2022estimation}.
Moreover, the contribution coming from these works is confined 
to a conceptual level, i.e., authors verified the performance of the vision algorithm, but the device was never tested in an active modality on the users \cite{qian2022predictive}.

\begin{figure*}[htbp]
\vspace{0.2cm}
    \centering
   \includegraphics[width=0.98\textwidth]{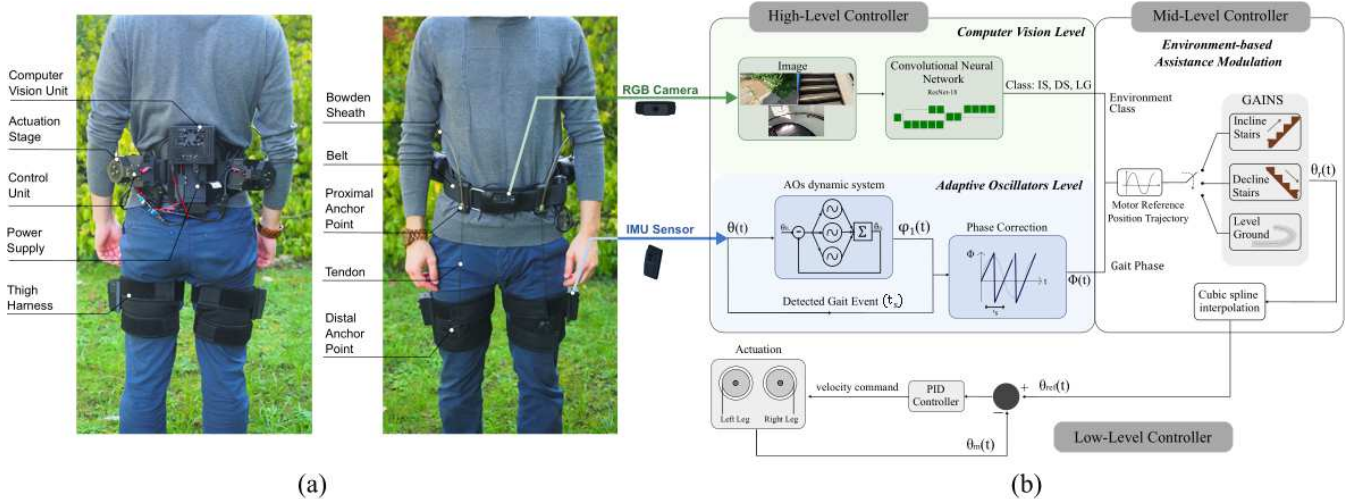}
   \vspace{-0.3cm}
    \caption{\textit{Hip exosuit design and real-time control framework} (a) The exosuit is a fully-actuated tendon-driven system that supports hip flexion during walking. It is made of a lightweight belt hosting the actuation stages, control unit, and batteries, and two thigh harnesses. Two inertial sensors capture the user's kinematics and an RGB camera scans the environment around the user (b) The Real-Time control framework is designed to distinguish among three kinds of terrains: Incline Stairs, Level Ground, or Decline Stairs. In the High-Level controller, images from the camera are processed in the \textit{Computer Vision level} in parallel to the users' gait phase estimation via Adaptive Oscillators (\textit{Adaptive Oscillators level}). The combination of these output is processed at the Mid-Level to generate an appropriate motor reference trajectory delivered to the actuator at the Low-Level.
   } 
    \label{fig:design}
\end{figure*}

In this framework, we developed and tested a novel control strategy for a hip exosuit incorporating computer vision to perform assistance modulation based on classification of the environment in front of the wearer recorded through an RGB camera.
Our controller is able to distinguish if the user is about to climb up or down a staircase, or to walk on level ground, and to adapt the assistance profile accordingly in order to comply with the change in the gait kinematic pattern \cite{riener2002stair}.
The underlying hypothesis is that an adaptive controller would improve the metabolic benefit of receiving assistance from the exosuit than a static controller (i.e., assistance magnitude does not change depending on the task).
To prove the point, we present here a comparison between the cases in which computer vision is active versus when it is not, to verify the enhanced performance of the system by using the former. 
As far as we are aware of, this is the first work in which a lower-limb exosuit incorporating computer vision is tested with active assistance complemented by an analysis of its effects on study participants.
Within the context of the present work, our main goal was to develop an assistive technology able to produce environment-based smooth and flexible assistance, such that users could take full advantage from its use and real-world deployment can be promoted in the long term.


\section{Exosuit Design}

The control strategy based on computer vision was tested on a fully-actuated exosuit to support hip flexion during natural human locomotion.
The assistive device (Fig.\ref{fig:design}-a) comprises of (i) two actuation stages, one for each leg, and main core of the active support system, (ii) a pair of textile harnesses wrapped around the users' thighs, (iii) and a belt (RDX Sport, Stafford, Texas, USA) used to hold the actuation and electronics.
Assistive forces are transmitted from the actuation stages to the user via external tendons (Black Braided Kevlar Fiber, KT5703-06, \SI{2.2}{\kilo\newton} max load, Loma Linda CA, USA).
The weight of the device is \SI{3.6}{\kilo\gram{}}, and it is powered by a battery (Tattu, \SI{14.8}{\volt}, \SI{3700}{\milli\ampere{}\hour}, 45C) allowing the user to receive active support for $\approx$ 10h of continuous operation.

Each actuation stage comprises of a flat brushless motor (T-Motor, AK60-6, \SI{24}{\volt}, 6:1 planetary gear-head reduction, Cube Mars actuator, TMOTOR, Nanchang, Jiangxi, China) driving a pulley with \diameter \SI{35}{\milli\meter{}} diameter wounding the artificial tendon of the corresponding leg.
Two Bowden cables (Shimano SLR, \diameter \SI{5}{\milli\meter{}}, Sakai, Ōsaka, Japan) connect the actuation units to two proximal anchor points, placed at the level of the waist and anchored to the belt, and are needed to cover the artificial tendons in charge of transferring the mechanical power of the motors to the user.
The artificial tendons are connected to the subject’s thighs via 3D printed distal anchor points sewed on the two soft fabric harnesses.

The sensors embedded in the suit comprise of two Inertial Measurement Units (IMU, Bosch, BNO055, Gerlingen, Germany), placed laterally on each thigh harness and used to detect the hip joint kinematics, each communicating with one control unit via Bluetooth Low Energy communication protocol (BLE, Feather nRF52 Bluefruit, Adafruit).

The control unit is in charge of running the real-time control algorithm: the \textit{High-Level Controller} incorporating the computer vision framework runs on an embedded processor (NVIDIA Jetson Nano, Santa Clara, CA, USA) at \SI{30}{\hertz} using an RGB camera (Logitech C920s PRO HD WEBCAM, Newark, CA, USA) placed in the frontal part of the belt as acquisition unit, while the \textit{Mid-Level} and \textit{Low-Level} run on microcontrollers (Arduino MKR 1010 WiFi, Arduino, Ivrea, Italy) at \SI{100}{\hertz}. The boards communicate with each other via Universal Serial Bus protocol.

The control framework was implemented in MATLAB/Simulink (MathWorks, Natick, Massachusetts MA, USA), with the exception of the Computer Vision algorithm written in Python.

\begin{figure*}[htbp]
\vspace{0.2cm}
    \centering
   \includegraphics[width=0.95\textwidth]{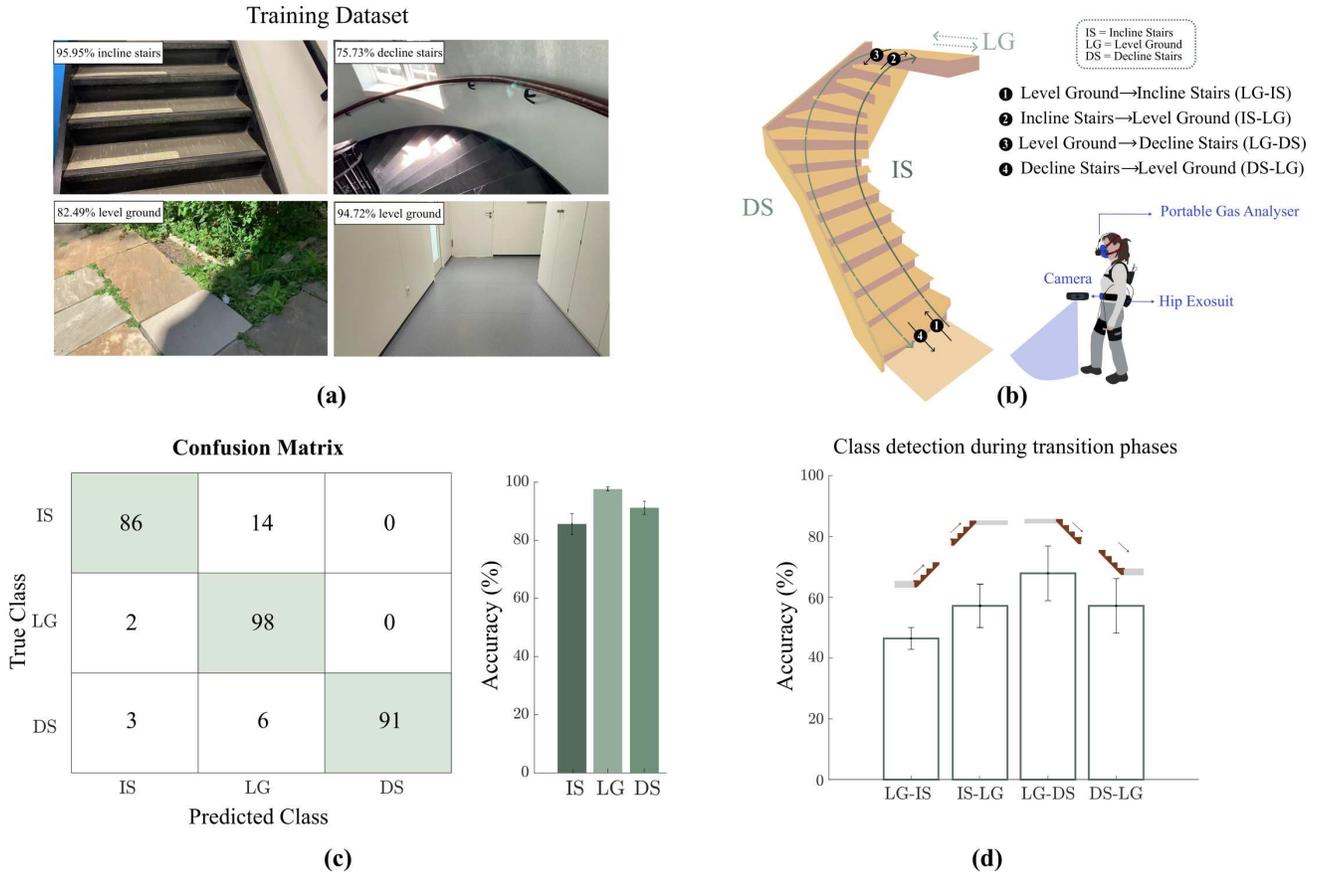}
    \caption{\textit{Computer Vision} (a) Examples of images contained in the training dataset for the ResNet-18 neural network. (b) Experimental Protocol:  participants walked throughout a path transitioning from level ground - stair climbing - level ground and viceversa. The task was repeated with the exosuit disabled, without Computer Vision enabled, and with Computer Vision. (c) Classification accuracy per class: each class reported a classification accuracy above $85\%$. The confusion matrix is shown to report the distribution of miclassifications across subjects. (d) Accuracy across transition phases: each bar represents the mean accuracy computed considering the step before and after each path transition. Here classification performance drops down to values $<80\%$.
   } 
    \label{fig:control}
\end{figure*}


\section{Real-Time Control Framework}
\label{sec:control}
We implemented a real-time Control Framework for our hip exosuit based on computer vision to determine the path in front of the user and to discriminate among three different walking conditions (i.e., incline stairs (IS), level ground (LG), and decline stairs (DS)) with the aim to perform an environment-based assistance modulation. \par

The control strategy (see Figure \ref{fig:design}-b) is composed of three layers: a \textit{High-Level Controller} running the computer vision algorithm in parallel to a gait phase estimation approach, a \textit{Mid-Level Controller} generating the actuator reference motion based on the environment and user's motion predicted at the High-Level, and a \textit{Low-Level Controller} implemented as a classical PID.

\subsection{High-Level Controller: \textit{Environment classification through computer vision and gait phase estimation}}
The High-Level Controller operates on two separate levels running in parallel to perform: (1) environment classification and (2) user's gait phase recognition through Adaptive Oscillators. 
We will refer to them as the \textit{Computer Vision level} and the \textit{Adaptive Oscillators level}.

\subsubsection{Computer Vision level}
The implemented Computer Vision level is based on a Convolutional Neural Network (CNN) model running at 30 Hz, able to predict the environment in the near space of the user through an RGB camera.
When it comes to image classification, CNNs have shown capabilities to outperform other kinds of classifiers, such as support vector machines \cite{laschowski2021computer}.
This is the case because multi-layer neural networks are able to optimally learn image features from training data, thus performing a more effective classification \cite{laschowski2022environment}.
However, they require a vast dataset for network training in order to promote generalization.

In our application, we adopted a CNN model composed of 18 layers (ResNet-18 \cite{he2016deep}) that was pre-trained on 10.500 images (see Section \ref{training}).
This net has been widely used in image classification applications for its low computational costs, high accuracy and simplicity to train \cite{he2016deep}.
The net takes as input the environmental images recorded in real-time by the camera placed on the belt of the user.
These are processed with a resolution of $400$x$400$ pixels and classified according to three classes: Incline Stairs (IS), Level Ground (LG), and Decline Stairs (DS).
Finally, the outputs of this level are the predicted class (i.e., IS, LG, and DS) and its confidence, which are sent, together with gait phase data, to the \textit{Mid-Level Controller}, as explained in a dedicated section (\ref{mid_level}).


\subsubsection{Adaptive Oscillators level}
The user's gait phase during locomotion is estimated in real-time from the hip joint kinematic (obtained by signals recorded by two IMUs) through the algorithm based on Adaptive Oscillators (\textit{AOs}) presented in our previous works  \cite{tricomi2021underactuated,zhang2022enhancing}.
The hip flexion angle of each leg $\theta(t)$ is reconstructed as sum of 3 sinusoids and indicated as $\hat{\theta}(t)$: \par

\begin{align}
&\hat{\theta}(t) = \alpha_{0}(t) + \sum_{n=1}^3 \alpha_{n}(t) \sin(\varphi_{n}(t))
\end{align}
whose amplitude $\alpha_{n}(t)$, phase $\varphi_{n}(t)$, frequency $\omega(t)$, and offset $\alpha_{0}(t)$ are continuously estimated via an error-driven approach via the function $F(t)=\theta(t)-\hat{\theta}(t)$, as follows:

\begin{align}
&\dot{\alpha}_{0}(t)=\eta F(t)\\
&\dot{\alpha}_{n}(t)=\eta F(t)\sin(\varphi_{n}(t))\\
&\dot{\varphi}_{n}(t)=\omega(t)\cdot n+\nu_{\varphi}\frac{F(t)}{\sum{\alpha_{n}}}\cos(\varphi_{n}(t))\\
&\dot{\omega}(t)=\nu_{\omega}\frac{F(t)}{\sum{\alpha_{n}}}\cos(\varphi_{1}(t))
\end{align}
Gains $\eta$, $\nu_{\varphi}$, and $\nu_{\omega}$ were set respectively equal to $5$, $20$ and $20$ from preliminary trials aimed at studying the convergence performance of the algorithm.

\begin{figure*}[htbp]
\vspace{0.2cm}
    \centering
   \includegraphics[width=0.9\textwidth]{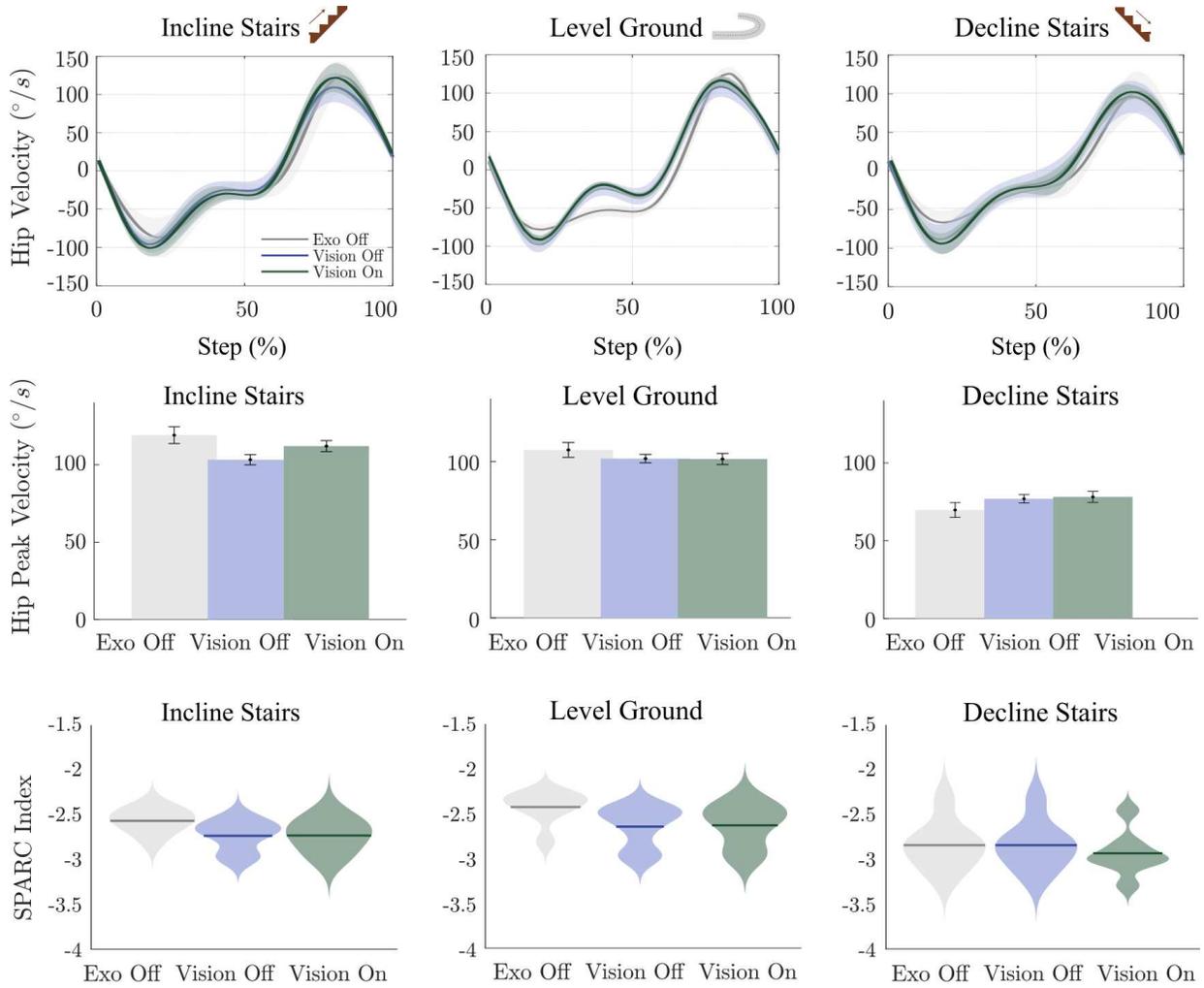}
    \caption{\textit{Kinematic Assessments}. Hip velocity profiles as percentage of the step duration for a typical subject are shown in the top row for the three path segments.
    Solid lines indicate the mean across steps and legs (grey for \textit{Exo Off}, blue for \textit{Vision Off}, and green for \textit{Vision On}); shaded areas represent the standard deviation.
    The second and third row report the peak velocity during swing as average across steps and subjects, and the movement smoothness expressed in terms of SPectral ARC length (SPARC) index.
   } 
    \label{fig:kin}
\end{figure*}

Among the AOs extracted quantities, the fundamental phase $\varphi_{1}(t)$ (i.e., $\varphi_{n}(t)$ with $n = 1$) is the one guiding the estimation of the user's gait phase:
$\varphi_{1}(t)$ is normalized in the range [$0$ $2\pi$) \SI{}{\radian} (i.e., $\varphi_{1}'(t) = \bmod(\varphi_{1}(t), 2\pi)$), and then shifted according to a corrective factor $\psi(t)$ in charge of aligning the beginning of each step with a value of $0$ \SI{}{\radian}:
\begin{align}
&\psi(t) =  \int{E_\psi\omega(t)e^{-\omega(t)(t-t_{s})}}
\end{align}
with $t_{s}$ being the instant in time indicating the beginning of each step and $E_\psi$ defined as below:
\begin{equation}
E_\psi = 
\begin{cases}-\varphi_{1}'(t_s) - \psi(t_{s}), &\mbox{ if } 0 \leq \varphi_{1}'(t_s) < \pi\\
2\pi-\varphi_{1}'(t_s) - \psi(t_{s}), &\mbox{ if } \pi \leq \varphi_{1}'(t_s) < 2\pi
\end{cases}
\label{eq:qdprg}
\end{equation}

The final gait phase variable, here named $\phi(t)$, is an indication of the progression along the gait cycle at each step, and it is eventually obtained as:
\begin{align}
\phi(t) = \bmod(\varphi_{1}'(t) + \psi(t), 2\pi)
\end{align}

\subsection{Mid-Level Controller: Environment-based assistance modulation}
\label{mid_level}
The Mid-Level Controller adjusts the level of assistance delivered to each leg according to the classification of the environment in front of the user performed by the \textit{Computer Vision level} and the estimated wearer's motion.

The reference motor position trajectory, $\theta_{r}(t)$, is generated from the gait phase data obtained from the \textit{Adaptive Oscillators level} as below:

\begin{equation}
\theta_{\mbox{\footnotesize{r}}}(t) =
\begin{cases}
M_{IS}\sin(\phi(t)-\pi) &\mbox{ if } \mbox{Incline Stairs}\\
M_{LG}\sin(\phi(t)-\pi) &\mbox{ if } \mbox{Level Ground}\\
M_{DS}\sin(\phi(t)-\pi) &\mbox{ if } \mbox{Decline Stairs}
\end{cases}
\label{ref_motion}
\end{equation}
being $M_{IS}$, $M_{LG}$, and $M_{DS}$ gains that modulates the amplitude of walking assistance depending on the environment classified at each step (IS, LG, and DS).
These three gains were tuned from preliminary trials according to kinematic considerations: we took into account the relative change between hip flexion profiles during swing according to the work of Riener et al. \cite{riener2002stair}.


Considering that environment classification is continuously performed throughout the walking according to the image acquisition frequency, in order to guarantee a smooth transition between assistive profiles, we implemented a "ranked-voting" procedure to decide the final class within each step performed by the user.
In detail, gait phase data are used to identify the times in which the subject reaches the maximum hip extension: this is, indeed, an indication of the beginning of the swing phase of walking and the time instant at which the device should apply zero assistive torque. In such a way, any change in the assistance amplitude does not cause users´s movement disruptions.
At this instant, which of the three equations in (\ref{ref_motion}) should take place is decided according to the overall classification confidence reached along the previous step:

\begin{equation}
\begin{cases}
C_{IS} = \sum_{n=1}^{N_{IS}} conf_{IS} & \forall \mbox{ $t_{n}$ with class IS}\\
C_{LG} = \sum_{n=1}^{N_{LG}} conf_{LG} & \forall  \mbox{ $t_{n}$ with class LG}\\
C_{DS} = \sum_{n=1}^{N_{DS}} conf_{DS} & \forall  \mbox{ $t_{n}$ with class DS}
\end{cases}
\end{equation}
where $C_{IS}$, $C_{LG}$, and $C_{DS}$ are counters indicating the goodness of prediction and are defined as the confidence values, $conf$, summed over the times that the respective class was predicted ($N_{IS}$, $N_{DS}$, $N_{LG}$) at each time stamp ($t_{n}$) along a step.
The class with the highest counter value is then chosen and the three counters are subsequently reset to zero.
The described procedure was implemented in order to enhance the robustness of the control algorithm to misclassifications occurring in the sporadic case of scarce lightening or confounding visual factors.


At last, the final actuator reference motion, $\theta_{ref}(t)$, is obtained from $\theta_{r}(t)$ after cubic spline interpolation and sent to the \textit{Low-Level Controller}.

\subsection{Low-Level Controller}
The Low-Level Controller compares the measured device state, i.e., actual position of the motor, with the desired one, i.e., the reference motion $\theta_{ref}(t)$, to minimize the error through reference tracking via closed-loop feedback position control.
The position error is converted into motor angular velocity through a PID controller.



\subsection{Computer Vision Training and Parameters Optimization}
\label{training}
The CNN chosen model (i.e., ResNet-18) was trained with a dataset composed of $5000$ manually selected images contained in the open-source dataset ExoNet \cite{laschowski2021computer},
complemented by $5500$ extra images acquired by the authors.
Figure \ref{fig:control}-a presents some examples contained in the training dataset.
The total $10500$ images were divided among the three classes and then scaled down to $400$x$400$ pixels. \par

The dataset was split such that the 80\% was used as training set, the 10\% as validation set, and the 10\% as test set. 
The network was trained for 60 epochs, using the Adam algorithm as optimizer, a batch size of $16$ and a learning rate of $0.1$, the latter decreased of ten times every 20 epochs. 



\section{Study Design}

Seven healthy subjects were enrolled to test the system ($3$ males/$4$ females, age $25.0\pm2.2$ years, mean$\pm$SD, weight $60.9\pm13.7$kg, height $169.0\pm9.9$cm). 
Informed consent forms were signed before the beginning of experiments. 
Research procedures were performed according to the Declaration of Helsinki and were approved by the Ethical Committee of Heidelberg University (resolution S-313/2020). 

\subsection{Experimental Protocol and Data Acquisition}
The study participants were asked to walk at their preferred speed along a path with total length equal to \SI{250}{\meter} arranged over different floors, comprising eight consecutive ramps of stairs (for a total amount of $88$ stairs, each $120$x$18$x$26$ \SI{}{\cm}) and \SI{150}{\meter} of level ground walking.
All subjects walked through the path starting from the bottom climbing up the full stairway from a level ground condition; once at the top they proceeded along the level ground path, and, finally, downstairs until the starting point (Figure \ref{fig:control}-b). \par

The path described above was repeated three times in three different conditions: (a) exosuit disabled (\textit{Exo Off}); (b) exosuit enabled without environment-based modulation (\textit{Vision Off}); (c) exosuit enabled with environment-based modulation (\textit{Vision On}).
Condition (b) - \textit{Vision Off} was implemented as described in section \ref{sec:control} by imposing the same gain to the three equations in (\ref{ref_motion}) (i.e., $M_{LG}$).
Participants rested for at least \SI{20}{\minute} in between trials to limit the effects of fatigue and conditions were randomized across subjects to avoid order effects.


\begin{figure*}[htbp]
\vspace{0.2cm}
    \centering
   \includegraphics[width=0.9\textwidth]{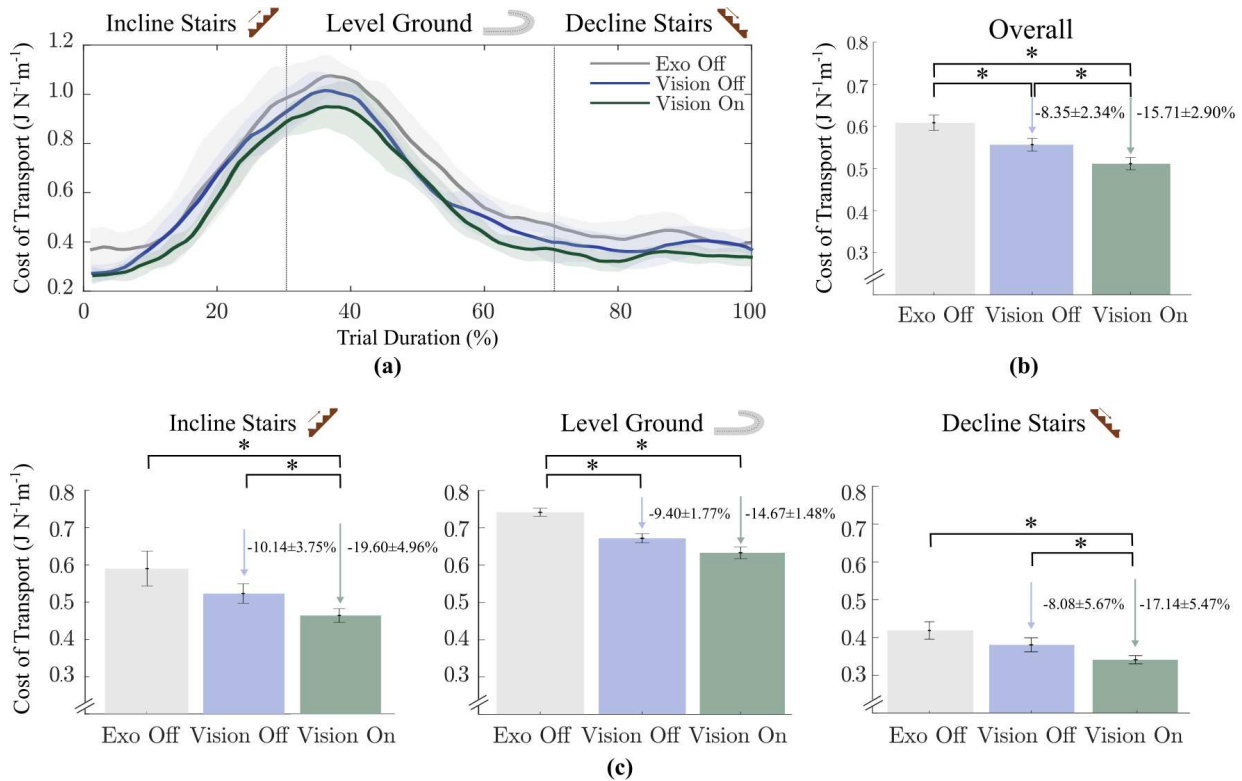}
    \caption{\textit{Cost of Transport}. (a) CoT profiles for the \textit{Exo Off} (grey), \textit{Vision Off} (blue), and \textit{Vision On} (green) conditions averaged across subjects.
    Solid lines represent the mean curve, while shaded areas are the standard deviations.
    For visual representation, timeseries are presented as percentage of trial duration.
    (b) Mean CoT computed on the overall path.
    (c) Mean CoT computed separately in the three path segments (IS, LG, DS).
    Savings for the \textit{Vision Off} and \textit{Vision On} conditions are computed in relation to \textit{Exo Off}. Statistical significance in pairwise comparisons is indicated with the symbol *.
   } 
    \label{fig:cot}
\end{figure*}

We collected oxygen and carbon dioxide consumption data through a portable gas analyser (K5, COSMED, Rome, Italy) to assess the metabolic energy expenditure and the effects of receiving assistance from the device with environment-based modulation compared to the remaining two conditions.
At the beginning of the experiment, subjects were asked to breath normally for \SI{4}{\min} in a standing resting condition.
The average metabolic cost at quite standing was subtracted from the metabolic data in later analysis to extract the cost of walking for each repetition of the task.

Classification data from the \textit{Computer Vision level}, the hip kinematics recorded from IMU sensors, and the actuator torques were also recorded for all subjects across repetitions.


\subsection{Data Analysis}

\textbf{Computer Vision}
We analysed the performance of the \textit{Computer Vision level} of the controller in terms of real-time class prediction accuracy. For each class (IS, LG, and DS) we assessed the number of correct predictions, True Positives (TP), as a percentage of the total number of predicted gait cycles belonging to a class, i.e., the sum of the TP and False Positives (FP):
\begin{equation}
   Accuracy (\%) = \frac{TP}{TP + FP}
\end{equation}
Moreover, to assess how this metric varies across transition phases, as the change from level ground to stairs and vice versa, we evaluated the accuracy of prediction corresponding to the step before and the one after each transition.
\vspace{0.1cm}
\par


\textbf{Kinematic Assessments}
Kinematic evaluations were performed on hip angular velocity profiles after segmentation into steps. 
We evaluated the effects of the \textit{Vision On} and \textit{Vision Off} conditions with respect to \textit{Exo Off} in terms of swing peak velocity and movement smoothness.
The latter was computed as Spectral ARC length (SPARC) index on the norm of the hip angular velocity as defined in \cite{balasubramanian2015analysis}.
\vspace{0.1cm}
\par

\textbf{Cost of Transport} Oxygen and carbon dioxide data were analysed to estimate the energy efficiency of walking for the \textit{Exo Off}, \textit{Vision Off}, and \textit{Vision On} conditions for the overall path and the three path segments (IS, LG, and DS).
Since subjects were instructed to complete the path at self-selected speed, to account for the different walking patterns across conditions, we analysed the metabolic data in terms of Cost of Transport (CoT) according to the following definition:
\begin{equation}
    CoT = \frac{P}{wgv}
\end{equation}
being $P$ the net metabolic cost of walking estimated using the Péronnet and Massicotte's equation \cite{peronnet1991table} subtracted by the average cost at quite standing, $w$ the subject's weight, $g$ the gravitational acceleration, and $v$ the average walking speed computed as covered distance over time. 

\subsection{Statistical Analysis}
We assessed the data normality distribution with the Shapiro-Wilk test ($\alpha = 0.05$).
All the analysed metrics resulted normally distributed.
We evaluated the three conditions presented in the study (\textit{Exo Off} vs. \textit{Vision Off} vs. \textit{Vision On}) for kinematics and CoT data via an one-way repeated measurements ANalysis Of VAriance (ANOVA).
Sphericity condition was evaluated using the Mauchly's test.
When the ANOVA results were significant, we performed a post-hoc analysis using paired t-tests with Bonferroni correction to evaluate pairwise differences between conditions. 
Reported measurements are presented as mean $~\pm$ standard error (SE).
\par

\section{Results}
\textbf{Computer Vision} The performance of the \textit{Computer Vision level} are presented in Figure \ref{fig:control}-c and -d in terms of accuracy per class and for transition phases as a mean across subjects and conditions.
For each class (Figure \ref{fig:control}-c), we obtained a prediction accuracy above $85 \%$: $85.56\pm3.58\%$ for IS, $97.57\pm0.69\%$ for LG, and $91.14\pm2.25\%$ for DS.
Results are graphically presented also in terms of confusion matrix to report the distribution of misclassifications.
Instead, lower values were obtained when analysing the classifier accuracy in correspondence of transition phases (Figure \ref{fig:control}-d): $46.43\pm3.57\%$ for the transition from LG to IS, $57.14\pm7.14\%$ from IS to LG, $67.86\pm8.99\%$ from LG to DS, and $57.14\pm8.99\%$ from DS to LG.
\vspace{0.1cm}
\par
 

\vspace{0.1cm}
\par

\textbf{Kinematic Assessments}
Results of the kinematic assessments are reported in Figure \ref{fig:kin} for the three path segments.
For the sake of a simpler result presentation, data are reported as mean between legs, since the one-way ANOVA showed no side-related difference.
The top row of Figure \ref{fig:kin} shows the time series of the hip angular velocity of a representative subject averaged across steps for the three conditions and divided by path segments.
Peak velocities reached during the swing phase and the movement smoothness are reported in the second and third row, respectively.
Statistical analysis did not show significant differences across conditions in any of the three path segments, as a sign of device transparency towards the users' physiological kinematic pattern.
\vspace{0.1cm}
\par

\textbf{Cost of Transport}
Results about metabolic evaluations are presented in Figure \ref{fig:cot}.
Exclusively for this analysis, one of the seven subjects was left out because of fatigue onset in subsequent tests.
Figure \ref{fig:cot}-a shows the Cost of Transport (CoT) timeseries as a percentage of trial duration averaged across subjects for the three conditions.
Considering the overall CoT along the path, results showed that subjects experienced the highest energy expenditure when walking in the \textit{Exo Off} condition, while the energy cost of walking decreased progressively during \textit{Vision Off} and \textit{Vision On}, with the latter reaching the lowest value (see Figure \ref{fig:cot}-a and b).
Here, CoT resulted significantly different both during \textit{Vision Off} (saving of $-8.35\pm2.34\%$, $p<0.05$) and \textit{Vision On} (saving of $-15.71\pm2.90\%$, $p<0.05$) with respect to \textit{Exo Off}.
The decreasing trend across conditions was confirmed by analysing the CoT separately in the three walking segments: we reached savings of $-10.14\pm3.75\%$ and $-19.60\pm4.96\%$ during IS for \textit{Vision Off} and \textit{Vision On} respectively, $-9.40\pm1.77\%$ and $-14.67\pm1.48\%$ during LG, and $-8.08\pm5.67\%$ and $-17.14\pm5.47\%$ during DS.
Considering the three walking segments, the \textit{Vision Off} condition showed a statistically significant difference than \textit{Exo Off} only for the LG walking ($p=0.004$), while \textit{Vision On} resulted always significantly different from \textit{Exo Off}  ($p<0.05$).
Additionally, in each presented case we found a statistically significant difference between \textit{Vision Off} and \textit{Vision On}, with the exception of Level Ground walking, where \textit{Vision On} showed slightly higher values than \textit{Vision Off}, yet not enough to achieve significance ($p=0.09$).

\section{Discussion and Conclusion}

By taking inspiration from the role that the visuomotor control plays in humans to promote adaptive locomotion, we developed a control strategy for a hip exosuit based on Computer Vision to scan the environment surrounding the user and to adapt the device assistance accordingly.

To the best of our knowledge, in this context, no other existing works in literature have reached our same advanced stage so far.
Works from Laschowski et al. \cite{laschowski2021computer, laschowski2022environment} are actively contributing to this research stream by providing open source databases of various walking environments in order to encourage the development of vision-based assistive systems for the lower-limbs.
Whereas, the study from Qian et al. \cite{qian2022predictive} is currently one of the most advanced in the field: along similar lines, the authors proposed an High-Level controller based on the combination of computer vision for terrains classification and an AOs-based approach for gait phase estimation.
Yet, the Mid- and Low-Level controllers were described just at a conceptual level, and subjects never walked with the device with active assistance mode.

Instead, we propose here a very first example of a fully developed controller for human locomotion assistance via hip exosuit in combination with an experimental evaluation to assess its efficacy.
In our system, information from wearable sensors are coupled with the ones coming from environment classification for the purpose of performing assistance modulation based on the near extra-personal space of the user.
Compared to our previous work (Zhang et al. \cite{ zhang2022enhancing}), through the incorporation of computer vision running in parallel to the wearer's kinematic estimation, we were able to exploit our system on diverse terrains, such as climbing up and down stairs, while achieving enhanced metabolic energy savings.

Our results proved that the \textit{Computer Vision level} was able to properly classify the three chosen scenarios (IS, LG, and DS) reporting an overall accuracy above $85\%$ for each class.
However, class detection during transition phases, i.e., class prediction corresponding to the step before and after each environment change, showed a higher rate of misclassifications and prediction accuracy dropped down to values $<80\%$.
Whether this could depend to sensitivity of camera position or other confounding factors is not clear and future studies should try to enhance robustness across gait transition phases.
In our experiments, these misclassifications did not cause any loss of balance or stability nor affected metabolic results in the overall evaluation, nonetheless sort of perfect accuracy in distinguish among different classes is preferred to avoid inconveniences of such kind \cite{inkol2020assessing}.

Kinematic assessments revealed that users could benefit from the exosuit assistance without experiencing an alteration of the physiological motion.
This is a good indicator that our device, being soft and lightweight, does not hinder subjects' movements, as expected from literature \cite{xiloyannis2021soft}.

Analysing the users' energy expenditure along the path, the inclusion of computer vision information, namely \textit{Vision On} condition, clearly outperformed \textit{Vision Off}, fully complying with our initial hypothesis.
Both in the overall path and in the analysis of path segments, in the \textit{Vision On} condition subjects were able to achieve significantly higher metabolic savings than \textit{Exo Off}, with a peak of $\approx$ $-20\%$ when climbing up the staircase and $\approx$ $-16\%$ considering the overall path.
Furthemore, CoT in \textit{Vision On} demonstrated to provide a significantly higher benefit also than \textit{Vision Off}, except during level ground walking.
This is indeed an expected result, given that for this class the controller was set to behave the same.
Even so, a trend toward reduction is still noticeable.
We can speculate that, since the metabolic rate is largely influenced by the preceding actions, a higher reduction in the \textit{Vision On} condition in LG may result from the larger saving obtained in the previous segment (i.e., IS).
To further prove the benefit provided by the \textit{Computer Vision level}, one may notice that the absence of assistance modulation (\textit{Vision Off} condition) did not bring to substantial benefits while climbing up or down the stairs. 

Possible limitations of our work include the recognition of solely three cases of terrains and poorer classification performances during path transition phases.
Next studies should assess the proposed control strategy in more diverse outdoor evaluations with the inclusion of an extended pool of recruited subjects and a higher number of terrains cases, for instance ascending or descending a ramp, or walk along a curved path.
Additionally, further efforts should focus on understanding how to properly detect and assist gait at the edge of environment change.

As of now, the promising results bode well in view of an extended assessment and a possible exploitation of our assistive wearable exosuit in the real world.
Just like in humans, the integration of a visual feedback seems to encourage higher human-robot symbiosis, thus bringing to lower energy expenditures.
In the long term, such metabolic savings could be beneficial in the case of wellness applications in older adults populations by promoting higher walking cadences or prolonged walking sessions.

\section*{Acknowledgements}
The results presented here were obtained as part of the project HeiAge (P2019-01-003) by Carl Zeiss Foundation.
We would like to thank the Carl Zeiss Foundation for the 5-year funding of SMART-AGE (P2019-01-003;2021-2026).

\bibliographystyle{IEEEtran}
\bibliography{./Bibliography/lib2.bib}

\end{document}